\documentclass{sig-alternate-05-2015}

\usepackage{amsmath}
\usepackage{graphicx}
\usepackage{hyperref}
\hypersetup{colorlinks=true}
\usepackage{subcaption}
\usepackage{float}
\usepackage{array}
\usepackage{ragged2e}
\usepackage[export]{adjustbox}
\usepackage{subcaption}

\begin{document}
\CopyrightYear{2016} 
\setcopyright{acmcopyright}
\conferenceinfo{PETRA '16,}{June 29-July 01, 2016, Corfu Island, Greece}
\isbn{978-1-4503-4337-4/16/06}\acmPrice{\$15.00}
\doi{http://dx.doi.org/10.1145/2910674.2910716}

%

\title{Evaluation of Deep Learning based Pose Estimation for Sign Language Recognition}

%
%
%
%
%

\numberofauthors{3} 
%
\author{
%
%
\alignauthor
Srujana Gattupalli\\
       \affaddr{Department of Computer Science and Engineering}\\
       \affaddr{University of Texas at Arlington}\\
       \affaddr{Arlington, Texas, USA}\\
       \email{srujana.gattupalli@mavs\\.uta.edu}
\alignauthor
Amir Ghaderi\\
       \affaddr{Department of Computer Science and Engineering}\\
       \affaddr{University of Texas at Arlington}\\
       \affaddr{Arlington, Texas, USA}\\
       \email{amir.ghaderi@mavs.uta.edu}
\alignauthor
Vassilis Athitsos\\
       \affaddr{Department of Computer Science and Engineering}\\
       \affaddr{University of Texas at Arlington}\\
       \affaddr{Arlington, Texas, USA}\\
       \email{athitsos@uta.edu}
}


\maketitle

\begin{abstract}
  Human body pose estimation and hand detection are two important tasks for systems that perform computer vision-based sign language recognition(SLR). However, both tasks are challenging, especially when the input is color videos, with no depth information. Many algorithms have been proposed in the literature for these tasks, and some of the most successful recent algorithms are based on deep learning. In this paper, we introduce a dataset for human pose estimation for SLR domain. We evaluate the performance of two deep learning based pose estimation methods, by performing user-independent experiments on our dataset. We also perform transfer learning, and we obtain results that demonstrate that transfer learning can improve pose estimation accuracy. The dataset and results from these methods can create a useful baseline for future works.
\end{abstract}

%
%

%
%

%
%
\printccsdesc



\section{Introduction}
Sign language communication is multi-modal. Some information is conveyed via manual features, such as hand motion and hand shape. Another information channel consists of facial features, such as lip movement, eye gaze, and facial expressions. A third channel is body posture, which can add to the meaning of a sign, or indicate change of subject in dialogues or stories. 

As pointed out in \cite{SLR:CooperHB11}, not much work has been done to utilize the non-manual feature body posture, which can be beneficial to SLR to aid recognition of certain positional signs e.g.``bruise'' or ``tattoo''  that involve pointing to or performing the movement on a certain body location. Moreover also pointed out in \cite{SLR:CooperHB11}, is that body posture can also be useful to differentiate between sign language dialogues and stories by observing changes in body positions while addressing different people. 

Our work on evaluation of human pose estimation techniques is aimed to contribute towards American Sign Language(ASL) recognition and hence we focus on localization of only upper body joints. Deep learning methods have found a lot of success in recent years on achieving good performance for classification problems as well as for localization and detection. The main benefit of using convolutional neural networks (CNNs) to address our problem is that they do not require features to be input from the programmer and are therefore less prone to human errors with regards to selecting appropriate features. Also, CNNs are holistic and take the entire image as input and are hence able to capture certain context that can be too complex to be performed by conventional technologies. Faster and accurate applications of CNNs can be successfully implemented with the help of GPUs and large amounts of available data. For our ASL recognition domain, we do not have large amount of data, so we improve performance accuracy of the deep learning methods using a technique called transfer learning. 

In this study, we introduce an RGB ASL image dataset (ASLID) and then investigate deep learning approaches on ASLID by performing body posture recognition by estimating positions of key upper body joint locations. Our ultimate goal is perform sign language recognition by efficiently obtaining upper body pose estimates over long video sequences and be able to perform multi-view pose estimation aided by this monocular pose estimation. The paper is furthermore organized as follows: In Section 2 we discuss related work; Section 3 we introduce our dataset ASLID; Section 4 transfer learning is explain and how we apply it towards pose estimation; In Section 5 we explain our experimental setup, training details and the evaluation protocol. Results for some existing methods evaluated in this paper are given in Section 6.

\begin{figure*}[!t]

\centering
  \includegraphics[height=3.5in,  keepaspectratio]{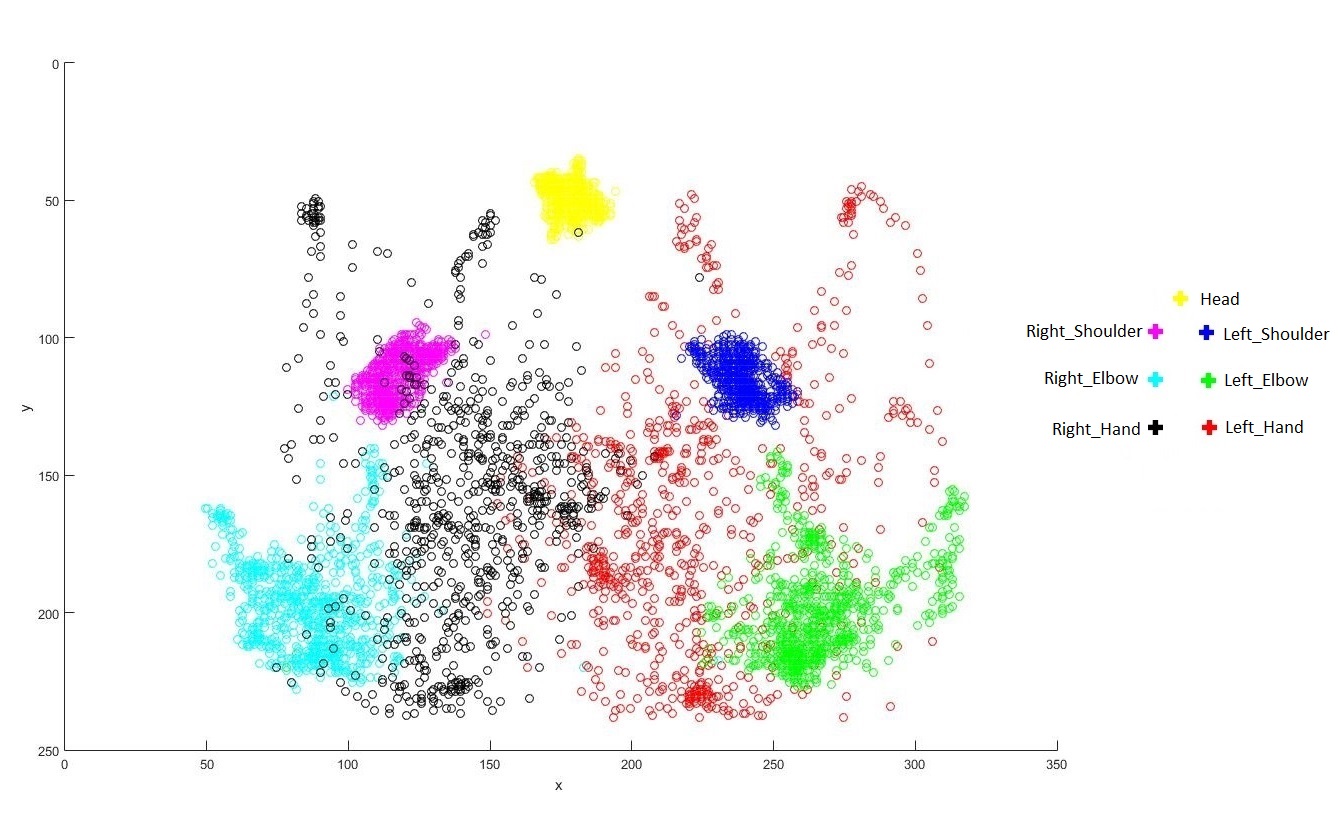}
  \caption{Ground truth training dataset variance}
  \label{fig:training_set_variance}
\end{figure*}
 \begin{figure*}[!t]
\centering
\begin{tabular}{c@{\hskip 12pt}c}
\begin{subfigure}{0.4\textwidth}\centering\includegraphics[height=1.8in]{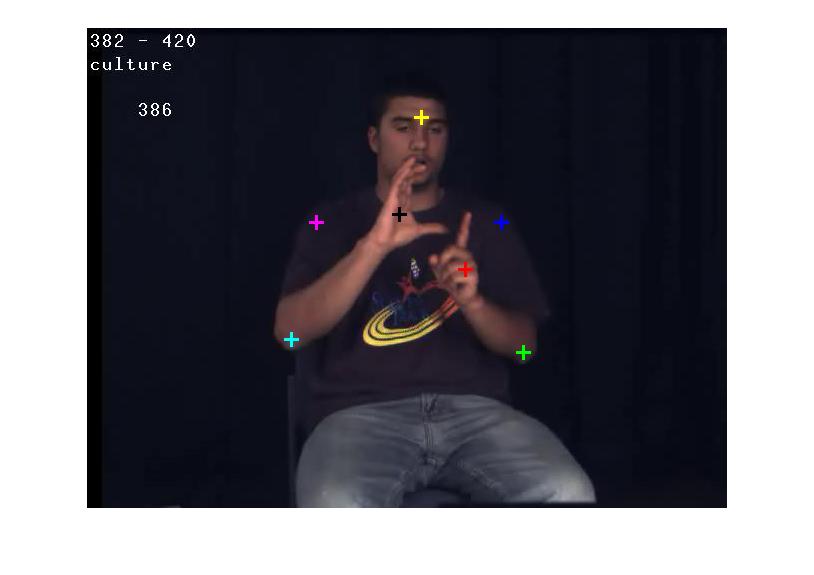}\end{subfigure}&
\begin{subfigure}{0.4\textwidth}\centering\includegraphics[height=1.8in]{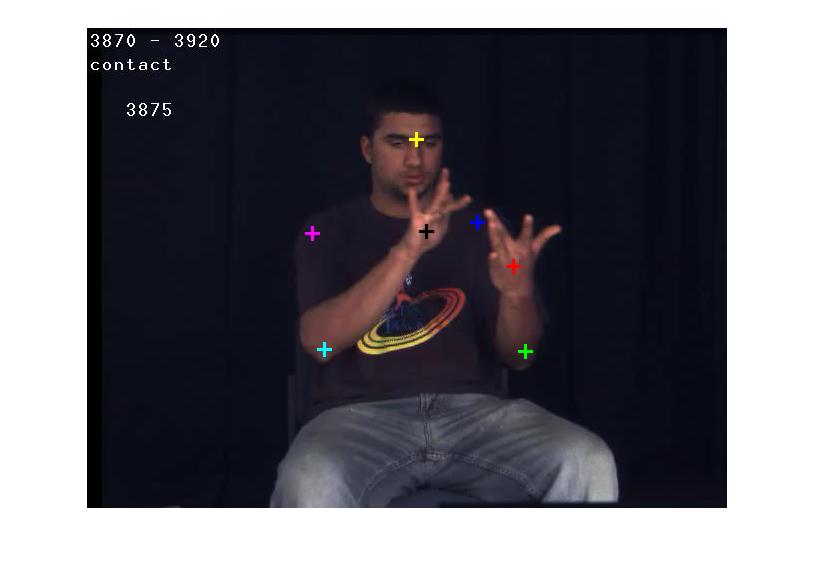}\end{subfigure}\\
\begin{subfigure}{0.4\textwidth}\centering\includegraphics[height=1.8in]{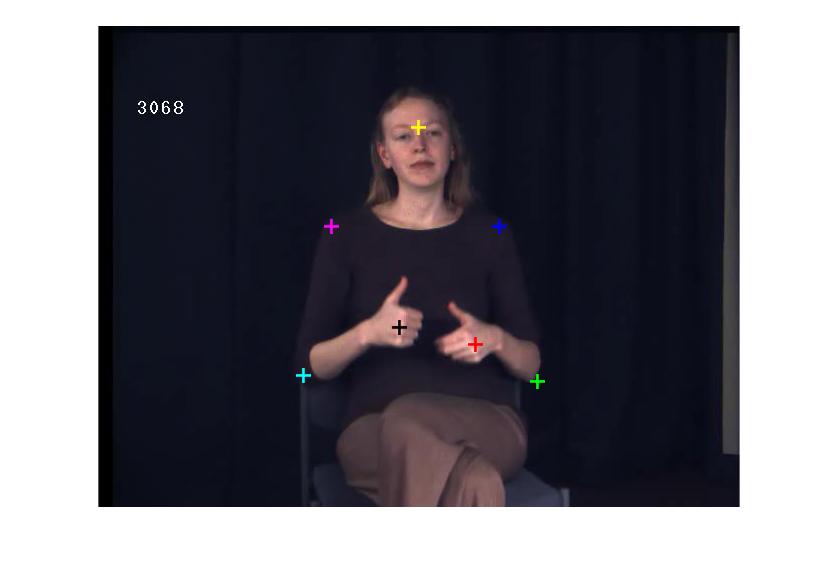}\end{subfigure}&
\begin{subfigure}{0.4\textwidth}\centering\includegraphics[height=1.8in]{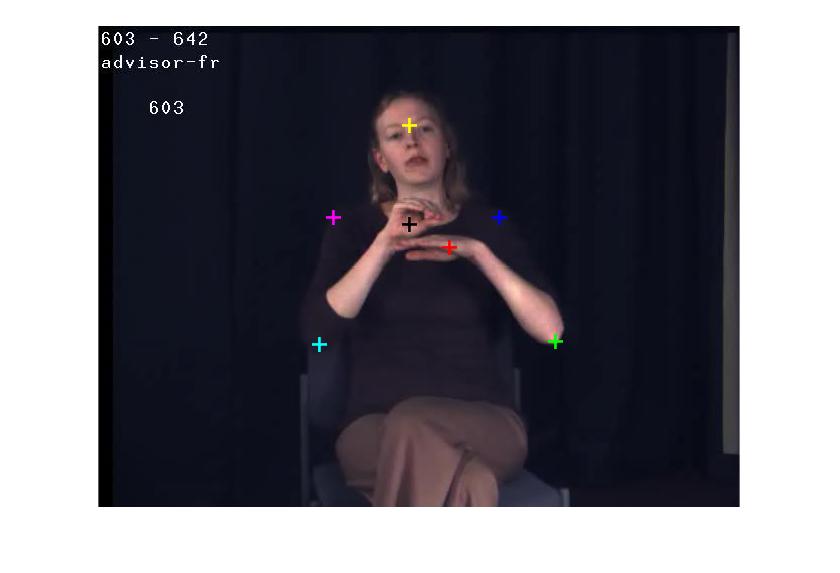}\end{subfigure}\\
\end{tabular}
\caption{Example image annotations from ASLID}
\label{fig:samples}
\end{figure*}
\section{Related Work}

Many techniques have been proposed for vision-based human pose estimation. Several  recent approaches use CNNs to address the task of joint localization for human body pose estimation. In \cite{t14nips}, Tompson et al. address the challenging problem of articulated human pose estimation in monocular images by capturing geometric relations between body parts, by first detecting the heatmaps of these body joint using a CNN architecture, and then applying a graphical model to validate these predictions. In \cite{Y11}, Yang and Ramanan create a tree-like model using local deformable joint parts and solve it using a linear SVM to achieve good results for pose estimation.  In \cite{articulated_xchen}, Chen and Yuille further extend this by using a graphical model and employing Deep CNNs to learn the conditional probabilities of presence of joints and their spatial relationships. Toshev and Szegedy \cite{deepposee} propose a deep learning based AlexNet-like CNN model for human pose estimation method by which they were able to localize body joints as solution to a regression problem, and then improve on the estimation precision by using a cascade of these pose regressors.

In \cite{tompson_1, tompson_2}, Jain, Tompson et al. perform pose estimation using a model that combines CNN to regress over joint heatmaps and a Markov Random Field(MRF). A relevant work on pose estimation specific to SLR domain is performed by \cite{AEH_zisserman} where they estimate joint locations over frames of sign language videos by first performing background subtraction and then predicting joints as a regression problem solved using random forest. The work by Pfister et al.\cite{Pfister15a} is also relevant to our SLR domain. In that work, the authors use a deep CNN to regress over heatmap of body joints, and improve performance by the use of temporal information from consecutive frames.

\begin{figure}
  \includegraphics[height=2.3in, width=3.5in]{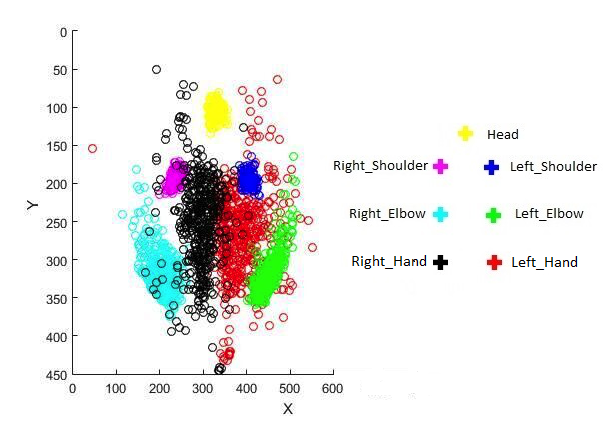}
  \caption{Ground truth test dataset variance}
  \label{fig:test_set_variance}
\end{figure}

\section{ASLID Dataset}
We present an American Sign Language Image Dataset (ASLID), with images extracted from Gallaudet Dictionary Videos\cite{Gallaudet} and the American Sign Language Lexicon Video Dataset (ASLLVD)\cite{ASLLVD:AthitsosNSNSYT08}. We provide annotations for upper body joint locations to perform body joint recognition. We have divided our dataset into training and testing sets, to help conduct user-independent experiments. Our training set consists of 808 ASLID images from different signs, performed by six different ASL signers. For the test set we have 479 ASLID images from two ASL signers from ASLLVD videos. The training and testing sets vary in terms of different users, signs and different colored backgrounds. We provide annotations for seven key upper body joint locations, namely left hand(LH), left elbow(LE), left shoulder(LS), head(H), right hand(RH), right shoulder(RS), right hand(RH). 

\begin{figure}[h]
  \includegraphics[height=2.5in, width=3.5in]{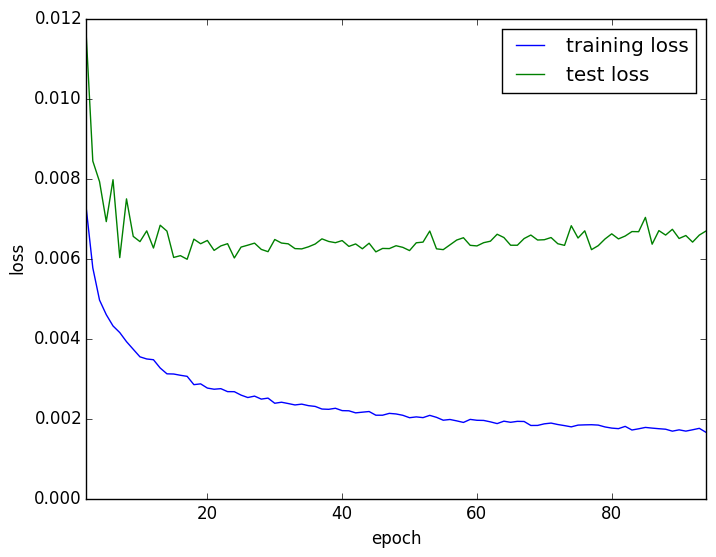}
  \caption{Loss for training and testing for Method(1)}
  \label{fig:log}
\end{figure}

Figure \ref{fig:samples} shows examples of annotated images from ASLID. Variations in the range of training and testing poses are shown in the ground truth scatter plots figures \ref{fig:training_set_variance} and \ref{fig:test_set_variance}. 

\begin{figure*}[!ht]
\centering
\begin{tabular}{c@{\hskip 12pt}c}
\begin{subfigure}{0.4\textwidth}\centering\includegraphics[height=1.8in]{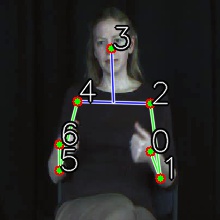}\end{subfigure}&
\begin{subfigure}{0.4\textwidth}\centering\includegraphics[height=1.8in]{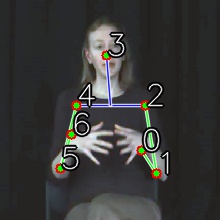}\end{subfigure}\\\\
\begin{subfigure}{0.4\textwidth}\centering\includegraphics[height=1.8in]{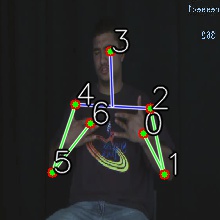}\end{subfigure}&
\begin{subfigure}{0.4\textwidth}\centering\includegraphics[height=1.8in]{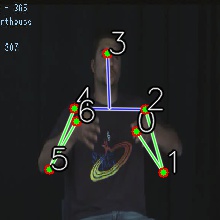}\end{subfigure}\\
\end{tabular}
\caption{Example visualizations of Pose Estimation by Method(1)}
\label{fig:result_deeppose_gallaudet}
\end{figure*}

\begin{figure*}[!ht]
\centering
\begin{tabular}{c@{\hskip 12pt}c}
\begin{subfigure}{0.45\textwidth}\centering\includegraphics[height=2.8in]{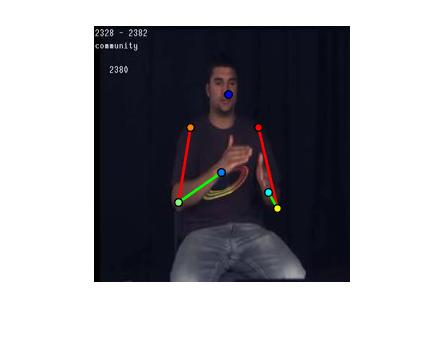}\end{subfigure}&
\begin{subfigure}{0.45\textwidth}\centering\includegraphics[height=2.8in]{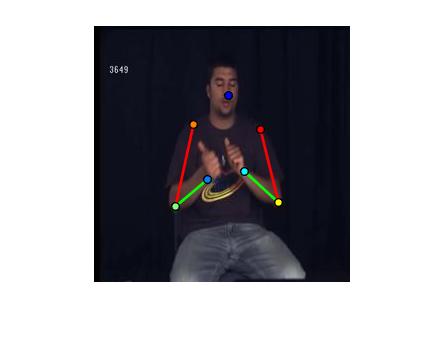}\end{subfigure}\\
\begin{subfigure}{0.45\textwidth}\centering\includegraphics[height=2.8in]{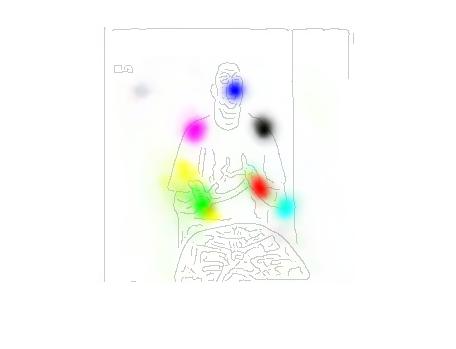}\end{subfigure}&
\begin{subfigure}{0.45\textwidth}\centering\includegraphics[height=2.8in]{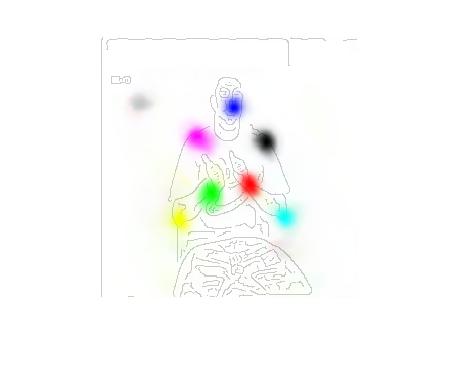}\end{subfigure}\\
\end{tabular}
\caption{Example visualizations of Pose Estimation by Method(5)}
\label{fig:result_pfister}
\end{figure*}

Our dataset and code to display annotations is available from:\\
\url{http://vlm1.uta.edu/~srujana/ASLID/ASL_Image_Dataset.html}. 

\section{Transfer Learning}
Transfer learning is a way to improve the performance of a learning algorithm by utilizing knowledge that is acquired from previously solved similar problem\cite{lu}. As pointed out in \cite{yosinski2014transferable}, initializing a network with transfer learned weights obtained even from a different task can improve performance compared to using random weights for initialization of a network. \cite{yosinski2014transferable} further points out that effectiveness of transfer learning is better if the difference between the original task and the target task is smaller. In our case, the original task is human body pose estimation and the target task is ASL specific upper body pose estimation, which are relatively similar. Hence, transfer learning helps in finetuning the pose estimator, so as to obtain better joint localization estimates for the ASL domain. In this paper we transfer the learned parameters from one method, to use as initial parameters before training using another method (See Table:1). In Method 3, we use the Deeppose network\cite{deepposee} to train on the FLIC dataset, and and transfer its learned parameters as initial weights for training on ASLID training set in Method 4. This shows significant improvement in performance, mainly attributed to the transferred learned weights. As pointed out in \cite{yosinski2014transferable}, this is also a way to avoid overfitting during training, even when we have a smaller target dataset than the original dataset.

\begin{table*}[!t]
\centering
\begin{center}
\begin{tabular}{|p{2cm}|p{3cm}|p{2.8cm}|p{1.5cm}|p{2cm}|p{2cm}|}
 \hline
 \centering\textbf{Method} & \centering\textbf{Training Model and Dataset} & \centering\textbf{Test Dataset} & \centering\textbf{Number of training images} & \centering\textbf{Number of test images} & \centering\textbf{Number of joints trained and tested} \tabularnewline [0.5ex] 
 \hline
 \centering
 1 &  \centering Method\cite{deepposee} trained on ASLID training set & \centering ASLID test set & \centering 808 & \centering 479 & \centering 7 \tabularnewline
 \hline 
 \centering
 2 & \centering Method\cite{deepposee} trained on Chalearn training set & \centering ASLID test set & \centering 433 & \centering 479 & \centering 7 \tabularnewline \hline  \centering
  3 & \centering Method\cite{deepposee} trained on FLIC dataset & \centering ASLID test set & \centering 17,378 & \centering 479 & \centering 7 \tabularnewline \hline \centering
    4 & \centering Method\cite{deepposee} trained on ASLID dataset started with weights from Method 3 & \centering ASLID test set & \centering 17,378 + 808 & \centering 479 & \centering 7 \tabularnewline \hline\centering
 5 & \centering Method\cite{Pfister15a} pre-trained on FLIC dataset & \centering ASLID test set & \centering 4.5K & \centering 479 & \centering 7 \tabularnewline \hline \centering
  6 & \centering Method\cite{deepposee} trained on ASLID training set(hands and face) & \centering ASLID test set & \centering 808 & \centering 479 & \centering 3 \tabularnewline
 [1ex] 
 \hline
\end{tabular}
\end{center}
\caption{Deep Learning based Pose Estimation Experiment Details}
\label{table:1}
\end{table*}

\section{Experiments}
 \justify{We evaluate performance of deep learning based pose estimators on static frames from ASL videos, by conducting user-independent experiments on images from the ASLID dataset. In this section we describe the experimental details and evaluation protocol. In the next section we show the results and present comparisons of methods. The caffe\cite{caffe} and chainer\cite{chainer_learningsys2015} frameworks were used for implementation. 
\\
\subsection{Pose Estimation Methods}
 The method proposed by Toshev et al.\cite{deepposee} uses deep neural networks for capturing context of body joints. We have trained the method mentioned in \cite{deepposee} on our ASLID training images and obtained results on our test set. We also use the model trained by Pfister\cite{Pfister15a} on FLIC dataset to obtain results on our dataset. In using the model of \cite{Pfister15a}, we only use the heatmap regression and spatial fusion parts of the method. We do not use optical flow, as we conduct pose estimation on static images, where flow information is not available. We compare results of ASLID pose estimation by models trained on other popular datasets (FLIC\cite{t14nips,FLIC} and Chalearn\cite{chalearn}) with the results by training on our dataset. Details of our experiments can be found in Table: \ref{table:1}.
\\
\subsection{Training Details}
 Toshev and Szegedy\cite{deepposee} proposed a deep learning based method, which localizes body joints by solving a regression problem, and further improves on estimation precision by using a cascade of these pose regressors. Their work demonstrates that a general deep learning based network originally formed for a classification problem can be fine-tuned and used to solve localization and detection problems. We have trained the model of Toshev et al.\cite{deepposee} on our dataset. We have performed user-independent evaluations using our ASLID training set and also on the benchmark Chalearn\cite{chalearn} and FLIC\cite{FLIC} datasets.
 
 In this paper we also use a pre-trained model on our dataset and compare its results with a model trained with our training set. Also this comparison creates a baseline for calculating improvement on results of other methods on the dataset. The pre-trained model which we have used for our evaluations is trained by Pfister\cite{Pfister15a}. The method by Pfister et.al. is interesting as they regress over heatmap of body joints instead of single centre co-ordinate of a body joint location, and they further improve performance by the use of temporal information from consecutive frames.

 We have trained the network for Methods 1 to 3 for 100 epochs, for the 7 key upper body joint locations. For Method 4, we have trained the network for 30 epochs and for Method 5 we use weights from Method 4 and train the model on ASLID training set for 10 epochs. The training and testing loss for Method 1 is shown in Figure-\ref{fig:log}. To improve accuracy of hand detection on ASLID, we performed experiments by training with DeepPose\cite{deepposee} only on the head and hand joint locations for 10 epochs (Method 6). 

\subsection{Evaluation Protocol}
We apply a quantitative evaluation measure similar to \cite{PAT:5540232} which they have applied for measuring hand detection accuracy. We extend this evaluation measure to be applicable for joint detection evaluation for seven upper body joint locations. The estimation is determined to be correct if the distance between detected joint and ground truth is less than a threshold($f$). The average face size of our test dataset is a reasonable threshold and we also calculate results using different threshold values.
Given a detected joint $i$ location estimate for an image $I$, $j_i(I)$, a ground truth joint location $g_i(I)$ and a threshold, $f$. We define our joint wise accuracy $Ai$ as:

\begin{equation}
 A_i(j_i(I),g_i(I),f) =
 \begin{cases} 
  1, & \text{if } \|j_i(I) - g_i(I)\| \leq f/2\\
  0,         	& \text{Otherwise}
  \end{cases}
\end{equation}
 \[\text{Where, } i \in \{LH, L	E, LS, H, RS, RE, RH\}\]
 
We display overall accuracy for corresponding joints as a mean of left and right joint detection accuracy for shoulders, elbows and hands. 

\section{Results}
To measure performance accuracy, we use the evaluation protocol as mentioned above. For methods 1 to 5, the average face width is 25 for the cropped and resized test dataset images. Our results for upper body joint locations are shown in graphs in Figures-\ref{fig:head},\ref{fig:hands},\ref{fig:shoulder},\ref{fig:elbow}. Example visualizations of some results on the ASLID test set are shown in Figures-\ref{fig:result_deeppose_gallaudet} and \ref{fig:result_pfister}. Here $f$ is the threshold value for measuring accuracy according to equation(1). The results demonstrate good pose estimation accuracy performance for sign language dataset.

In Method 4, we use the Deeppose network\cite{deepposee} to train on the FLIC dataset and transfer its learned weights as initial weights for training on ASLID. The results show huge improvement on Method 3 and demonstrate that transfer learning can help improve pose estimation performance of a method through the transferred knowledge from another trained model. 
In our experiments, we have also conducted training on 3 joint locations(H, LH, RH) to improve hand localization accuracy. We measuring accuracy for i=3 using equation(1). Here, the average face width is 50 for the cropped and resized test dataset images. The resultant accuracy achieved for hands and face is shown in Figure-\ref{fig:hands_only}.



We try six different methods on our dataset and show the comparison of results for experiments. As the figures show Method 4 and 5 work very well on detecting the body joints.  
\begin{figure}[p]
\centering
\includegraphics[height=2.5in]{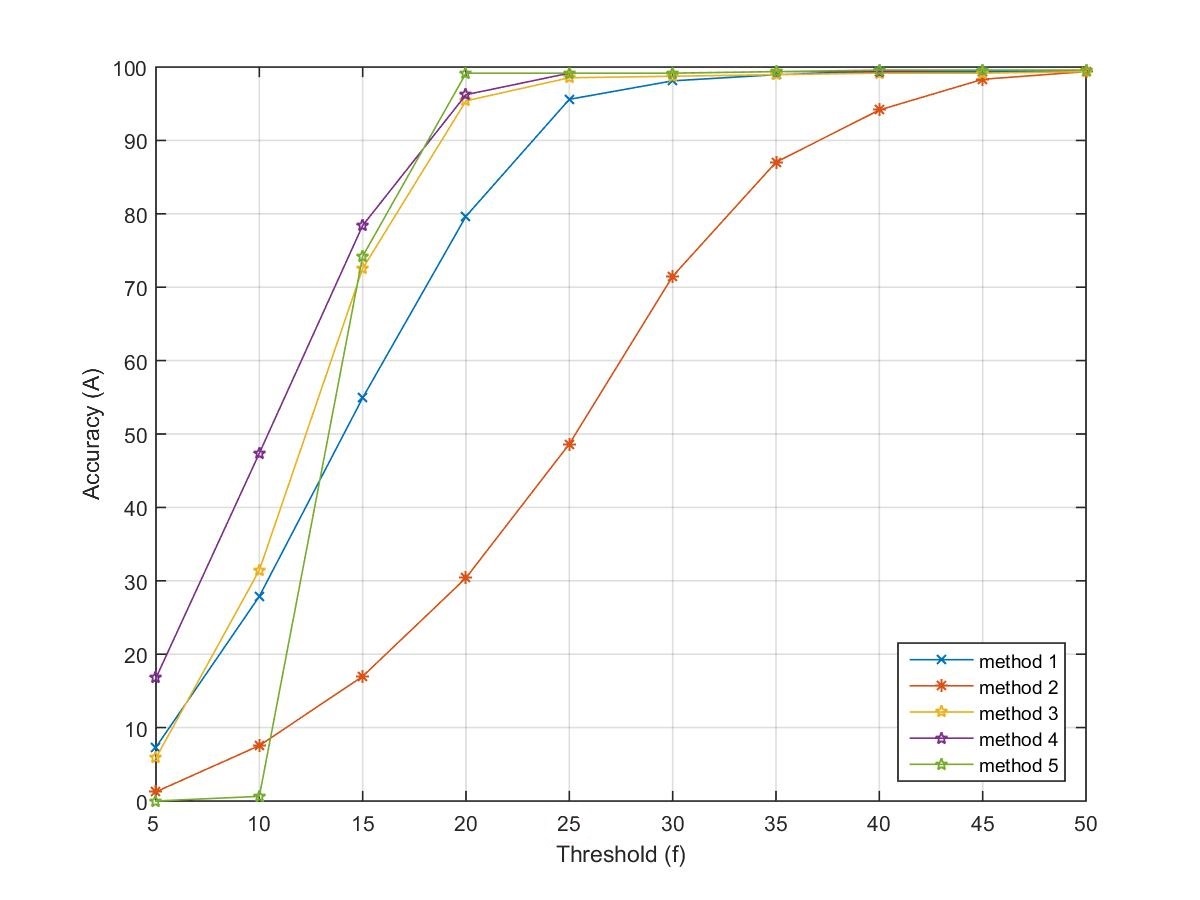}
\caption{Head detection results for Methods 1 to 5}
\label{fig:head}
\end{figure}

\begin{figure}[p]
\centering
\includegraphics[height=2.5in]{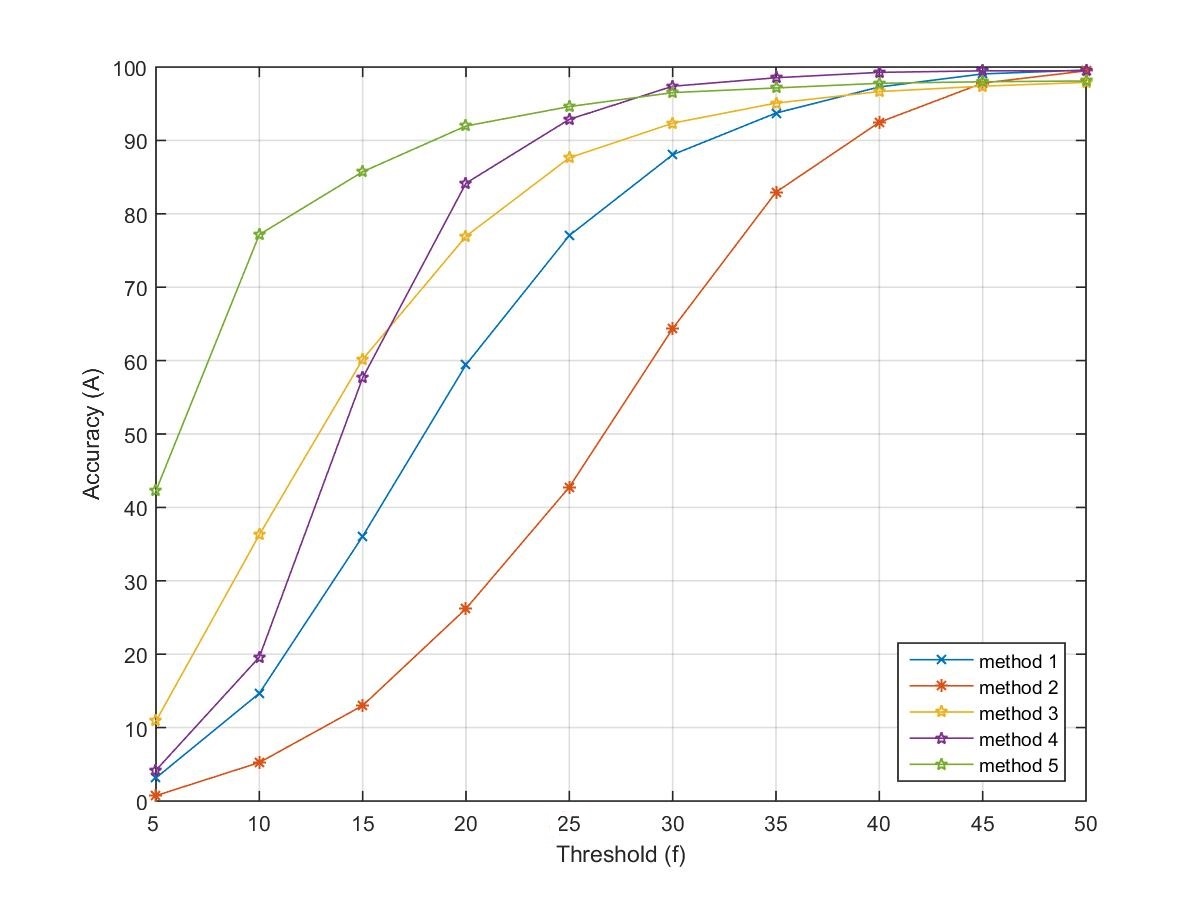}
\caption{Hand detection results for Method 1 to 5}
\label{fig:hands}
\end{figure}

\begin{figure}[p]
\centering
\includegraphics[height=2.5in]{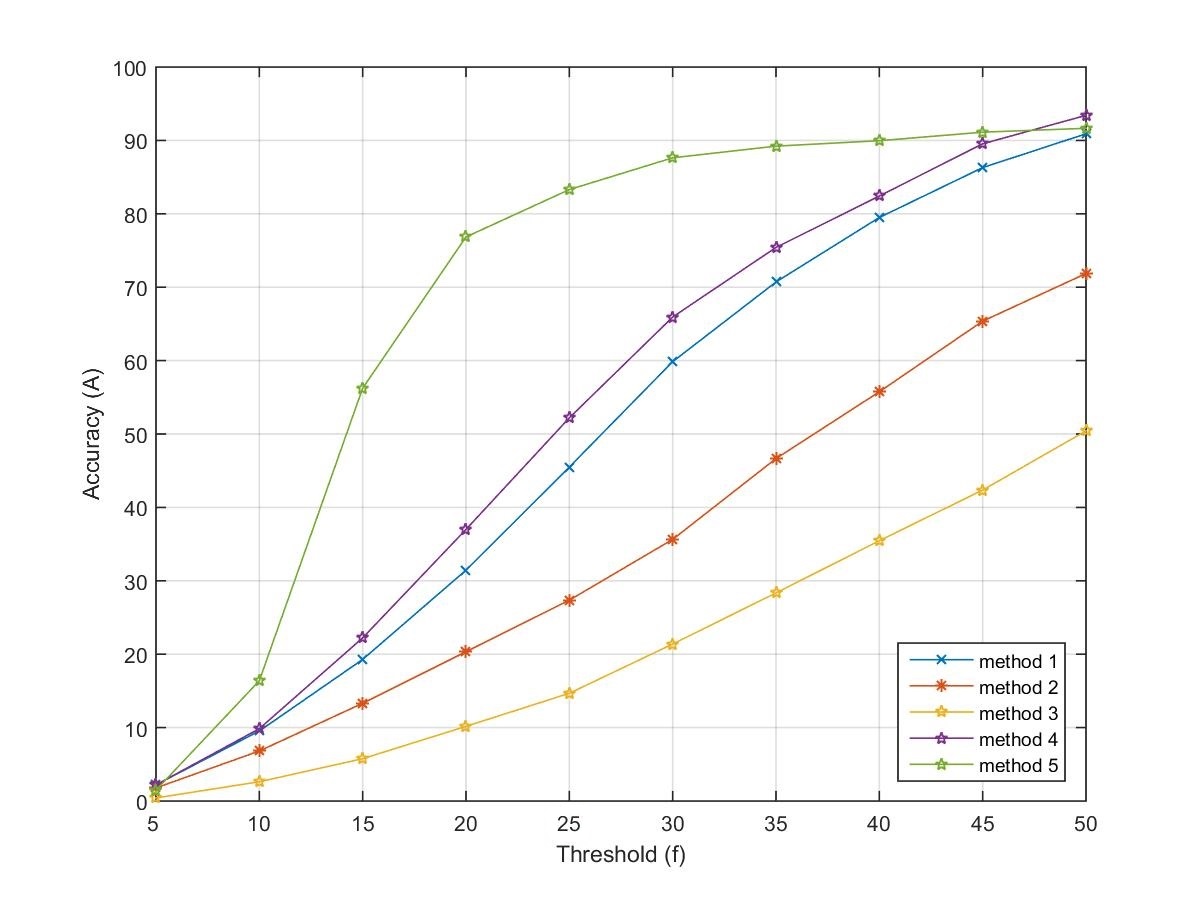}
\caption{Shoulder detection results for Method 1 to 5}
\label{fig:shoulder}
\end{figure}

\begin{figure}[t!]
\centering
\includegraphics[height=2.5in]{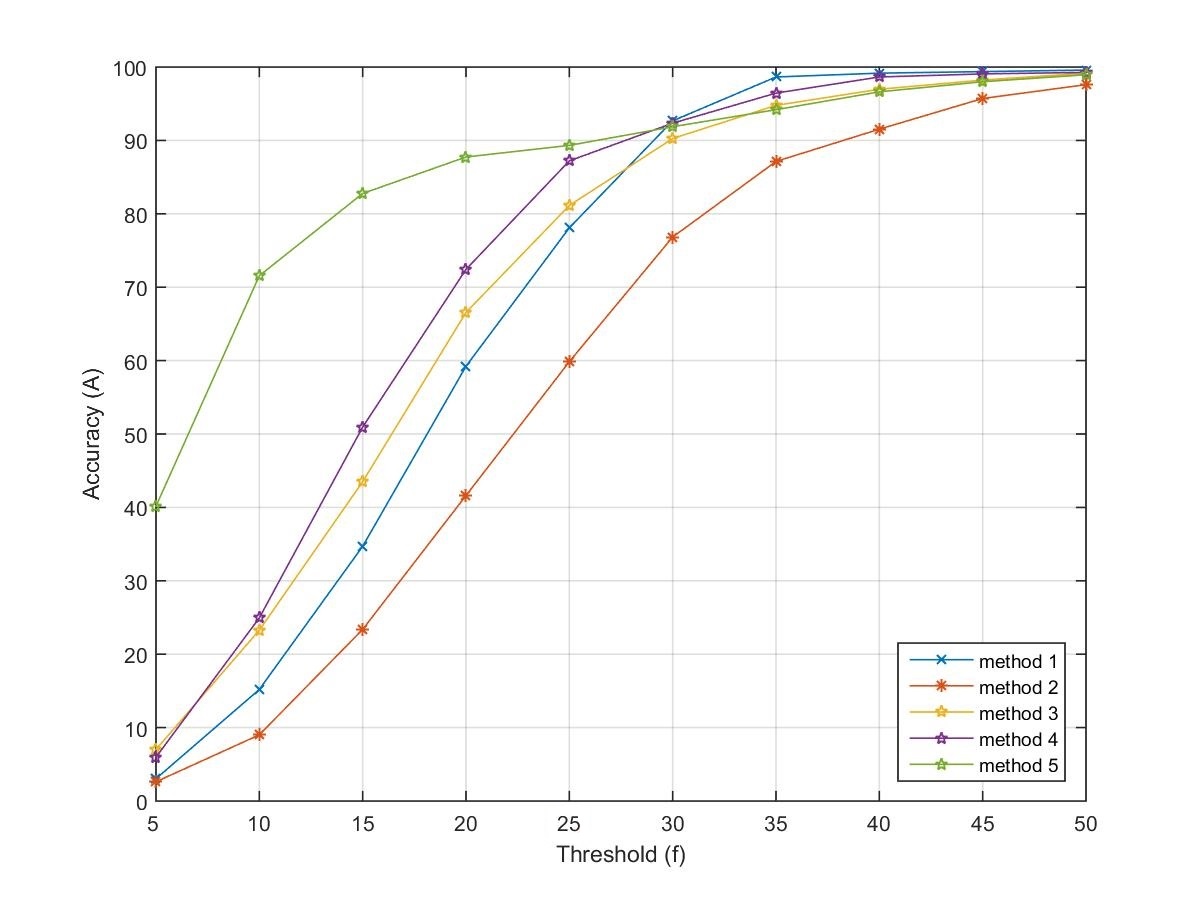}
\caption{Elbow detection results for Method 1 to 5}
\label{fig:elbow}
\end{figure}

\begin{figure}[t!]
\centering
\includegraphics[height=2.5in]{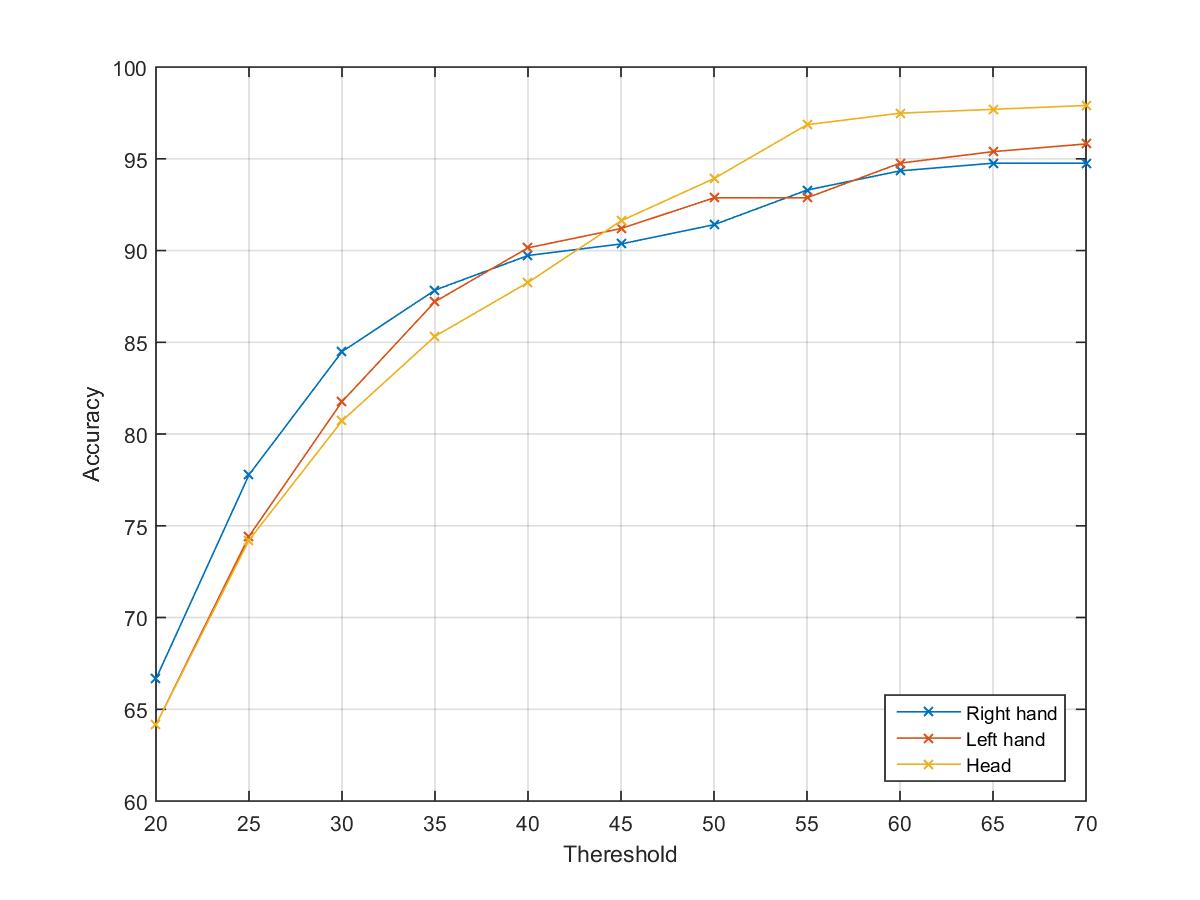}
\caption{Results for Method 6}
\label{fig:hands_only}
\end{figure}

\section{Conclusion and Future Work}
The work in this paper focuses on pose estimation in static images. An interesting direction is to extend the work to pose estimation on ASL videos, so as to provide useful features for ASL recognition. Our aim is to improve sign recognition accuracy, by performing motion analysis on features from an efficient automatic human pose tracker, that detects and tracks upper body joint positions over continuous sign language video sequences. 

In summary, this paper has presented a new image dataset for pose estimation, aimed towards applications in sign language recognition. This dataset can be used for measurement of performance of existing methods as well as methods to be developed in the future. In this paper we have selected two deep learning based state-of-the-art methods for human pose estimation, and we have measured the accuracy of them on our dataset. This measurement creates a baseline for other methods in this domain. 

\section{Acknowledgments}
This work was partially supported by National Science Foundation grants IIS-1055062 and CNS-1338118. Any opinions, findings, and conclusions or recommendations expressed in this publication are those of the authors, and do not necessarily reflect the views of the National Science Foundation.
 


%
\justify
\bibliographystyle{abbrv}
\bibliography{sigproc}  
%
%
 
\end{document}